\pdfoutput=1
\documentclass[runningheads]{llncs}

\usepackage{graphicx}
\usepackage{subfig}
\usepackage[symbol]{footmisc}
\usepackage{url}

\newcommand{\subf}[2]{%
  {\small\begin{tabular}[t]{@{}c@{}}
  #1\\#2
  \end{tabular}}%
}
\begin{document}
%
\title{An Autoencoder Based Approach to Simulate Sports Games}

%
%
\author{Ashwin Vaswani\thanks{Equal contribution} \inst{1}\and 
Rijul Ganguly$^{*}$\inst{1}\and 
Het Shah$^{*}$\inst{1}\and 
Sharan Ranjit S$^{*}$ \inst{1}\and
Shrey Pandit \inst{1} \and 
Samruddhi Bothara \inst{1} }
\authorrunning{A. Vaswani et al.}
%
\institute{Birla Institute of Technology and Science Pilani, K. K. Birla Goa Campus}

\maketitle

%
%

%
%
%

\begin{abstract}
 Sports data has become widely available in the recent past. With the improvement of machine learning techniques, there have been attempts to use sports data to analyze not only the outcome of individual games but also to improve insights and strategies. The outbreak of COVID-$19$ has interrupted sports leagues globally, giving rise to increasing questions and speculations about the outcome of this season’s leagues. What if the season was not interrupted and concluded normally? Which teams would end up winning trophies? Which players would perform the best? Which team would end their season on a high and which teams would fail to keep up with the pressure? We aim to tackle this problem and develop a solution. In this paper, we propose \textbf{UCLData}, which is a dataset containing detailed information of UEFA Champions League games played over the past six years. We also propose a novel autoencoder based machine learning pipeline that can come up with a story on how the rest of the season will pan out.

\end{abstract}
\section{Introduction}\label{sec:introduction}

Sports analytics has received extensive attention over the past few years. While a lot of work in sports analysis emphasizes on visual \cite{visual1,visual2} and tactical analysis \cite{tactical}, there have been recent attempts to predict the outcome of individual games and entire seasons. However, most of these attempts only predict the outcome without providing insights or internal statistics to corroborate their results. Another issue is the lack of large clean datasets for this task. While most of the existing datasets provide data summarising matches, there is little focus on the little intricacies of matches that might be of interest. To tackle this, our proposed \textbf{UCLData} dataset consists of both match and individual statistics from Champions League matches played over the past six years. Further, we handle dataset size issues with the help of some intuitive priors or handcrafted features which make our model robust and realistic. 

In this work, our proposed novel autoencoder based architecture not only predicts the outcome of a game but also predicts its internal statistics, to give a more holistic picture of how a match is expected to pan out. Moreover, apart from match-wise statistics, we also present player-wise statistics to provide details about the contribution of each player and minor details about a match which are generally ignored. The code for our work is made publicly available.\footnote[2]{ \textit{https://github.com/ashwinvaswani/whatif}}
\section{Related Work}\label{sec:relatedworks}

Most of the previous approaches based on machine learning for predicting results of sports games aim to predict simply the outcome of matches, instead of running a simulation predicting all match-related statistics.

Kampakis \textit{et al.}\cite{kampakis2015using} used both player and team data for cricket matches to predict the performance of teams based on different features. A study by Rotshtein \textit{et al.}\cite{geneticball} used several predictive models to predict outcomes in the English Premier League and the Premiership Rugby in England. There are various works based on Bayesian models \cite{footbayesian,kickoffai,ieee34}, but these limit themselves to predicting the outcomes of individual football matches instead of running simulations. A work based on the Gaussian Process model by L. Maystre \textit{et al.}\cite{gaussian} attempts to learn the strengths and traits of a team by player wise contributions. This is an inspiration for our present study. 

Huang \textit{et al.}\cite{worldcup} focus on using neural networks to predict the results of the $2006$ Football World Cup and this is the most similar to what we have tried to achieve in this paper. They achieved an accuracy of $76.9\%$ on the games' results, having special difficulty in predicting draws. Hucaljuk \textit{et al.}\cite{noconf} incorporated expert opinion into Champions League matches, but in this case, there was no increase in accuracy in their prediction of game scores. S. Mohammad Arabzad \textit{et al.} \cite{nn1} incorporated the use of neural networks for the Iranian premier league.  Flitman \textit{et al.} \cite{nn2} developed a model that will readily predict the winner of Australian Football League games together with the probability of that win. This model was developed using a genetically modified neural network to calculate the likely winner, combined with a linear program optimisation to determine the probability of that win occurring in the context of the tipping competition scoring regime.
\section{Dataset}\label{sec:dataset}
The following section details our approach for creating a dataset from which we can derive meaningful predictions.
\begin{figure}[ht]
    \centering
    \includegraphics[width=0.8 \textwidth, height=65mm]{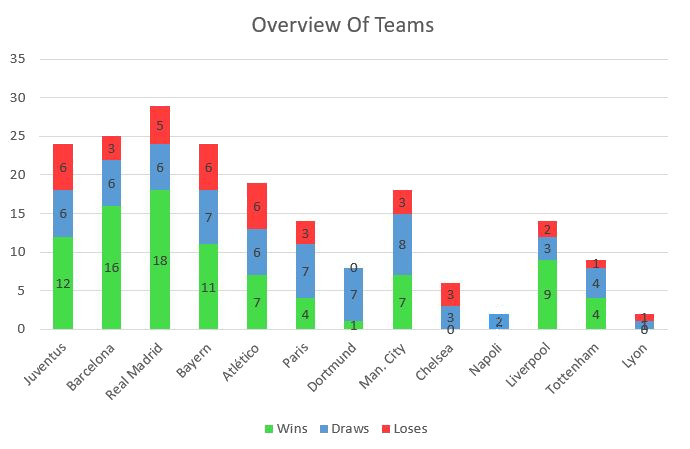}
    \caption{Overview of Dataset}
    \label{fig:dataset}
\end{figure}

\subsection{Data Collection}

We scrape data from the official UEFA Champions League website to build our dataset. Data from the years $2014$ to $2020$ is used. Overall we collect the data for $157$ knockout stage matches. We do not collect data for group stage matches because our predictions will be on the knockout stage games of the $2019-20$ season of the Champions League, and hence we did not want the context of group stage matches misleading our model.

To scrape the data, we use the Python library Beautiful Soup \cite{soup}, which assists us to take the data directly from the relevant websites. We divide our data into two categories - team data and player data. Team data contains the statistics for the entire team playing in the match on both sides, while player data includes the statistics of the teams' individual players. 

To obtain team data, we use the official UEFA website for the individual matches. However, the official website does not contain the statistics for individual players. Hence, we extract individual player data from the FBref website \cite{fbref} and the Global Sports Archive website \cite{gsa}. Table \ref{tab:stats} summarises the attributes we considered for our dataset. 
\begin{table}
    \centering
    \begin{tabular}{|l|l|}
    \hline
         & \hspace{5pt}Attributes \\
    \hline
     Team   & \hspace{5pt}Total goals, total attempts, attempts on and off target, blocked shots, \\ & \hspace{5pt}shots which hit the woodwork, corners, off-sides, amount of possession, \\ & \hspace{5pt}total passes, passing accuracy, completed passes, distance covered, \\ & \hspace{5pt}number of balls recovered, tackles, clearances, blocks, yellow and \\ & \hspace{5pt}red cards, fouls.\\
     \hline
     Individual \hspace{5pt}&\hspace{5pt}Goals scored, total shots, shots on target, assists, interceptions, crosses,\\ & \hspace{5pt}fouls committed, player off-sides, total time played\\
     \hline
    \end{tabular}
    \vspace{5pt}
    \caption{List of attributes for a team and an player}
    \label{tab:stats}
\end{table}

\subsection{Data Pre-processing}

Our data in its raw form contains numbers spanning a wide range - from hundreds in the fields such as passes completed to only one or two in areas such as goals. Passing such fields without any pre-processing would lead to our proposed model not accurately capturing this wide range. Hence we normalize our data to the range of zero to one using MinMax Scaling. This ensures that our model does not give any undue importance to any fields because of scaling issues. After pre-processing, we create embeddings from our normalized data.

\subsection{Creation of Embeddings}

There are some problems with using individual match data throughout. First, information from earlier matches cannot be used efficiently. This argument can be demonstrated with the help of an example. Let us say two teams A and B play against each other in years Y1 and Y2. Now, these two games are not independent as the two sides have played multiple other teams in this period and improved their game-play. Thus, it is not ideal to directly use individual match stats without capturing this context. Another issue is regarding players switching teams, which is quite common in sports. If a player plays in team A in year Y1 and switches to team B in year Y2, we need a way to represent it so that their individual information is maintained. We solve these problems with the use of embeddings. We create embeddings for each team and each player so that when two teams are matched up, these representations can capture the interactions with other teams and players and can preserve contextual information from previous data.   
\section{Methodology}\label{sec:method}

\subsection{Handling problem of Data bias}\label{subsec:databias}
Our data consists of matches from the last six years of Champions League games. Although we found this data sufficient to capture relationships between teams and players, there were a few issues due to imbalance. Some teams, not being Champions League regulars, had fewer data points. We find that our initial results were biased towards the lower number of data points of these teams and lacked generalization. We attempted to overcome this issue with the help of prior information, which is important in the field of football analysis. We propose three additional hand-crafted features which are crucial in the context of a game. We also infer that regularisation and dropout help in solving some of these problems. We show in the following sections how the addition of each of these features helps in making our results more robust. 

\subsubsection{Home / Away status:}\label{subsec:homeaway}
\begin{figure}[ht]
    \centering
    \subfloat[Home / Away wins \label{sub:1}]{\includegraphics[width=0.62 \textwidth, height = 45mm]{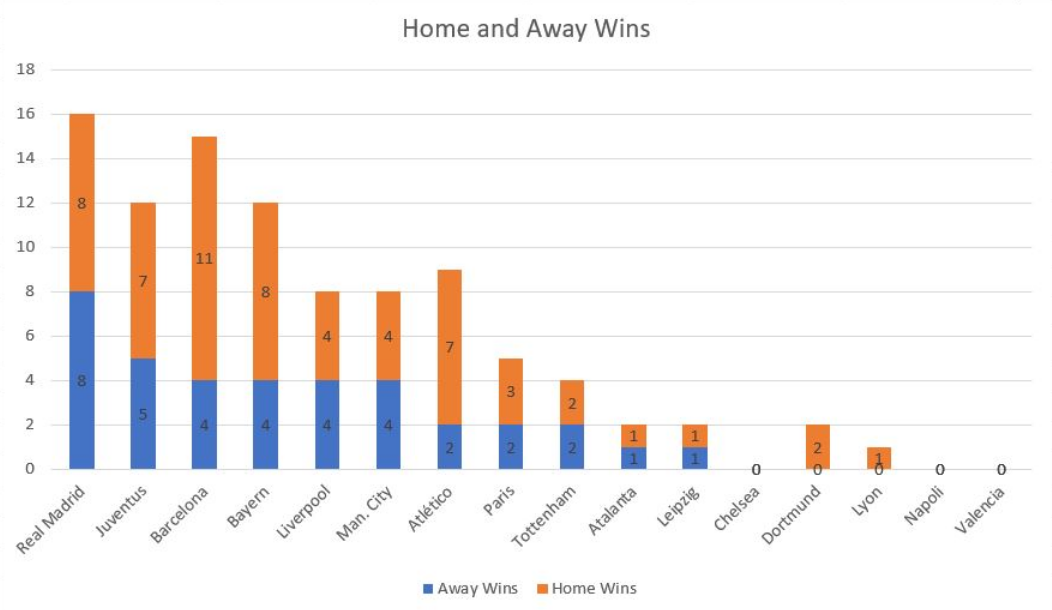}}
    \subfloat[Outcome vs Form - Colour intensity represents higher concentration of matches with a particular outcome. \label{sub:2}]{\includegraphics[width=0.38 \textwidth, height = 45mm]{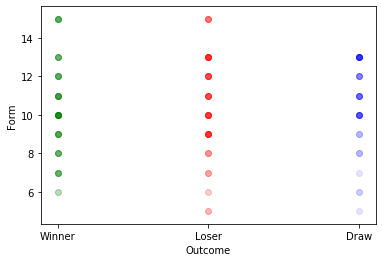}}
    \caption{Home / Away wins and Outcome vs Form}
    \label{fig:home}
\end{figure}

An important feature of Champions League knockout stages is the Home / Away concept. A fixture consists of two games wherein each game is played at the home ground of the two teams. The Figure  \ref{fig:home}\subref{sub:1} shows some analysis of the importance of the location of the fixture. 
It can be seen that there is a general trend for most teams to perform better at home than while away, which is quite intuitive. We attempt to use this information by adding an extra flag to indicate the team is playing at home apart from our embeddings while giving input to the model.

\subsubsection{Form Index:}\label{subsec:form}
Another essential feature, relevant to the context of a match, is the form of the two teams playing. It can be seen in Figure \ref{fig:home}\subref{sub:2} that at lower values of the form($<$ $7$), teams are less likely to win whereas, in the middle range, it's difficult to predict with just form. We used the recent results of each team (Results from the five most recent games before the fixture) to generate a form index by giving a score of three points to a Win, one to a Draw, and zero to a Loss.  This additional information helped in improving results of certain matches as a team would rather go into a game with a form of $15$(five straight wins) than $0$(five straight losses).

\subsubsection{Experience:}\label{subsec:experience}
Figure \ref{fig:dataset} shows that some teams such as Real Madrid, being Champions League regulars have plenty of data points. In contrast, teams like Atalanta, who are new to the Champions League, have few data points. Hence, results of matches involving Atalanta were biased to the data from these limited games resulting in Atalanta performing exceptionally well against the odds in our initial experiments. While this can be considered a case of an "upset" or Atalanta being "dark horses", we wanted to improve our results and make our predictions robust. A critical factor is a team's experience in the Champions League, due to the pressure of playing in such a high-profile platform. We accumulated total matches played by every team in our data to account for this experience factor, which helped in solving the issue of predictions being biased because of limited data.

\subsection{Details of the Model}\label{subsec:details}

\begin{figure}
\centering
\begin{tabular}{c}
\subfloat[Teams Model]{\includegraphics[width=0.99 \textwidth]{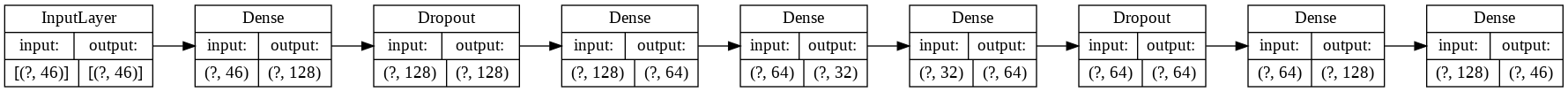}}
\\
\subfloat[Players Model]{\includegraphics[width=0.99 \textwidth]{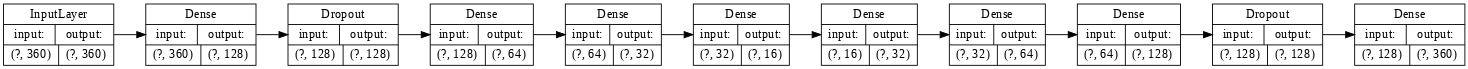}}
\\
\end{tabular}
\caption{Details of the models used}
\label{fig:models}
\end{figure}

\begin{figure}[ht]
    \centering
    \includegraphics[width=0.8 \textwidth, height=50mm]{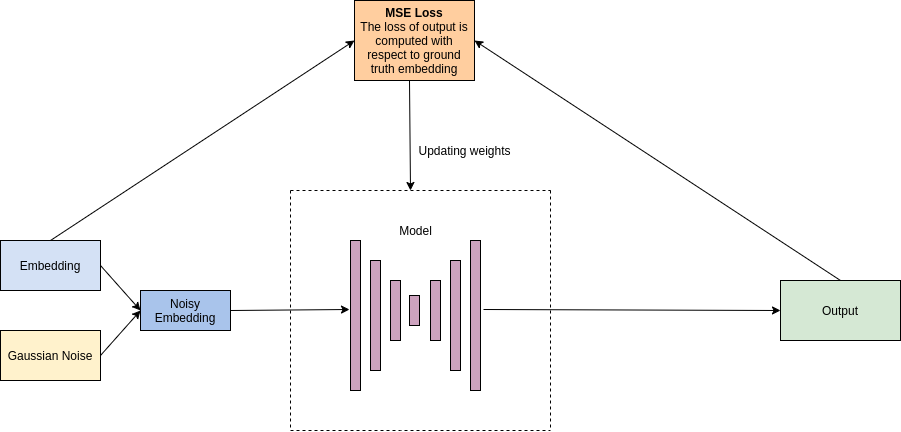}
    \caption{Summary of our pipeline}
    \label{fig:pipeline}
\end{figure}

Our network is based on the idea of autoencoders\cite{autoencoder} which are widely used for data compression. The aim of our training process is to learn about the various features of the team and the players. To achieve this we aim to learn an embedding in latent dimension. We also want this data in latent dimension to be robust from other factors which cannot always be predicted from the data. The model architectures are as shown in Figure \ref{fig:models}. We add a Gaussian noise to this in order to create a "noisy" embedding. This is given as an input to the network. The intuition for adding Gaussian noise is that it will help take into consideration some factors which are not consistent with the data (example a player having a lucky day or an off day/weather conditions which affect the play). We use the embedding without Gaussian noise as our ground truth labels. The schematic of the training process is given in the Figure \ref{fig:pipeline}. So, after the training process, the model learned some important insights about the team's/player's performance, which is later helpful during the playoffs to decide the winner of a particular match. For training, the loss is taken to be \textbf{mean squared error}, and the metric that we have considered is the \textbf{root mean squared error} (RMSE). We used Adam Optimizer with a learning rate of $0.01$ and the batch size was $10$ embeddings, for both our models. The RMSE values in the training and validation process are not metrics of performance of the model on new matches, rather they are indicators of the model’s efficiency in learning the embedding. The training RMSE value for the team model is $0.1380$, and for players model is $0.1127$. The validation RMSE values for both the models are pretty close to the training models at $0.1379$ for the team model and $0.1126$ for the players' model. The overall summary of our pipeline can be seen in Figure \ref{fig:pipeline}.

\section{Results and Observations}\label{sec:results}

\begin{figure}[ht]
    \centering
    \includegraphics[width=0.9 \textwidth, height=45mm]{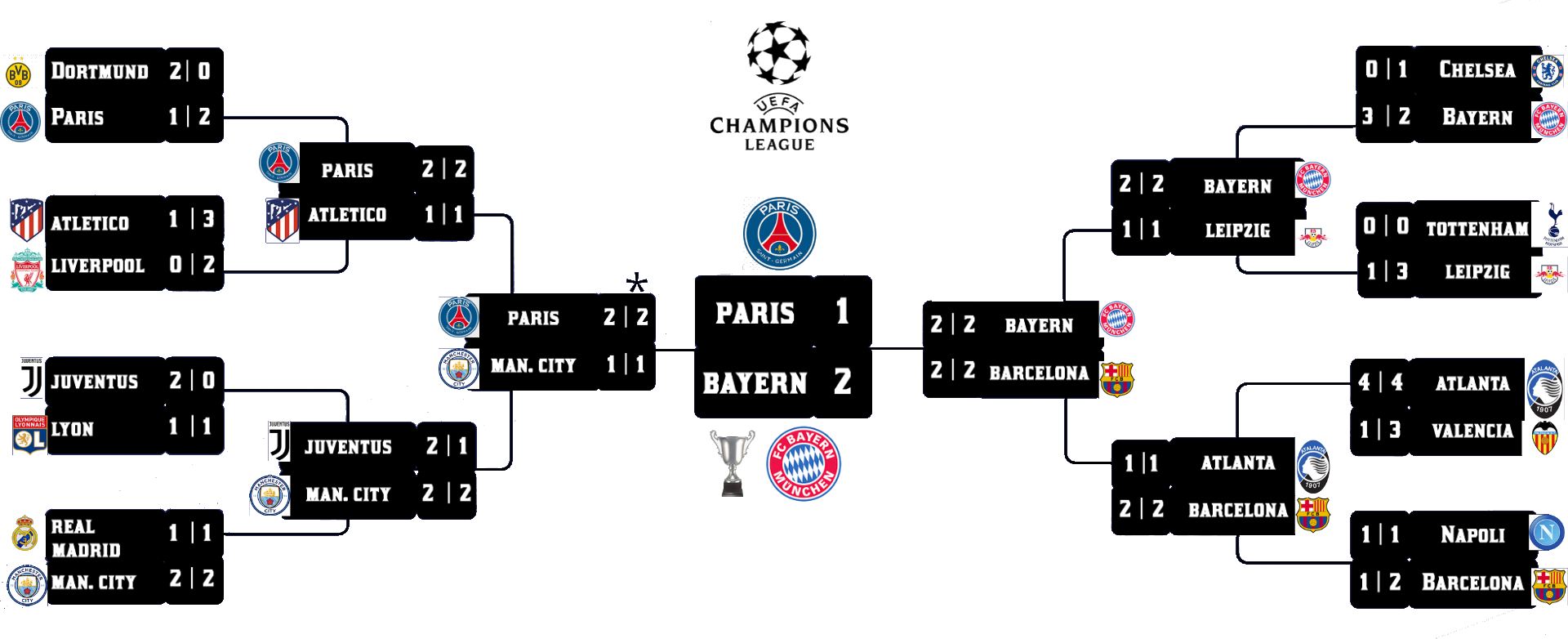}
    \caption{Overview of Simulation}
    \label{fig:sim}
\end{figure}

\begin{figure}[ht]
    \centering
    \subfloat[Correlation with goals \label{sub:cor}]{\includegraphics[width=0.5 \textwidth, height=50mm]{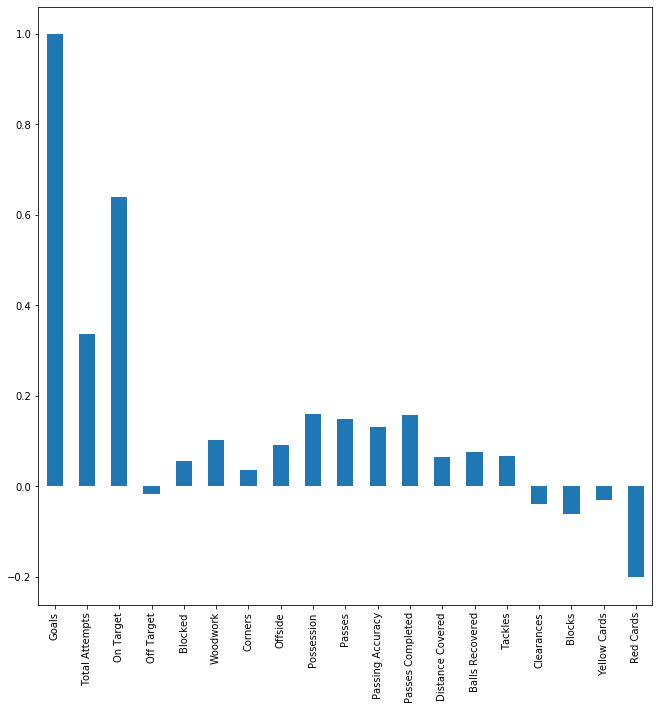}}
    \subfloat[Average Goals per team \label{sub:cross}]{\includegraphics[width=0.5 \textwidth, height=50mm]{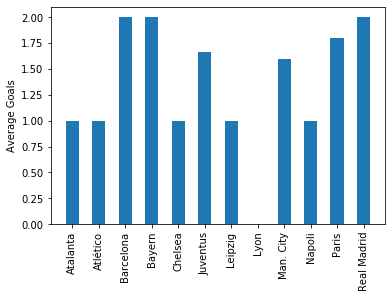}}
    \caption{(a) Shows high correlation between goals and shots on target (b) Identifies high / low scoring teams}
    \label{fig:corr}
\end{figure}

\begin{figure}[ht]
\centering
\begin{tabular}{c c c}
\subf{\includegraphics[width=0.32 \textwidth, height=40mm]{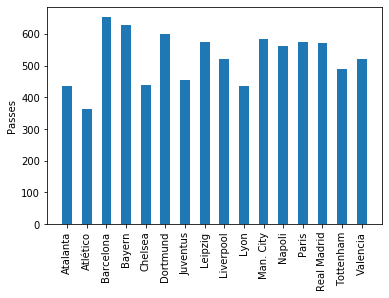}}
     {(1a) Passes}
&
\subf{\includegraphics[width=0.32 \textwidth, height=40mm]{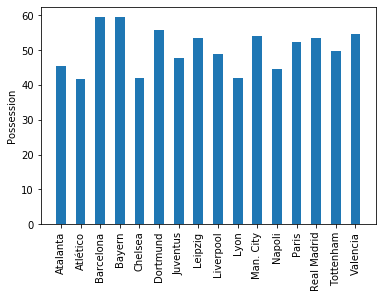}}
     {(2a) Possession}
     
&
\subf{\includegraphics[width=0.32 \textwidth, height=40mm]{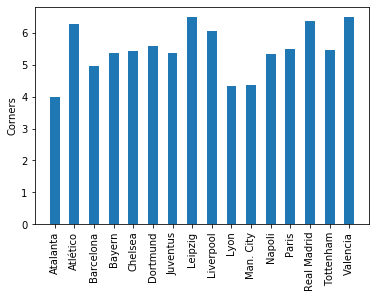}}
     {(3a) Corners}
\\
\subf{\includegraphics[width=0.32 \textwidth, height=40mm]{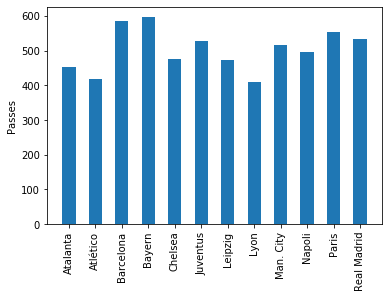}}
     {(1b) Simulation of Passes}
&
\subf{\includegraphics[width=0.32 \textwidth, height=40mm]{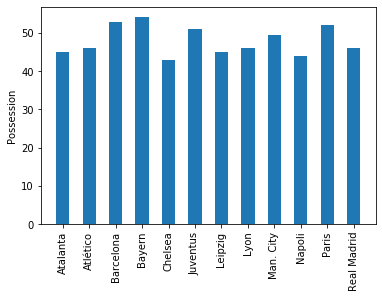}}
     {(2b) Simulation of Possession}
&
\subf{\includegraphics[width=0.32 \textwidth, height=40mm]{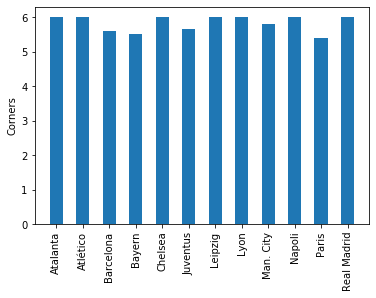}}
     {(3b) Simulation of Corners}
\\
\end{tabular}
\caption{Distribution of Passes, Possession and Corners in training data and in our simulations. The similarity between the plots show that our model is able to learn the distribution effectively. Figure (1a) and (1b) are Passes vs Teams, Figure (2a) and (2b) are Possession vs Teams and Figure (3a) and (3b) are Corners vs Teams}
\label{fig:stats}
\end{figure}

Figure \ref{fig:sim} gives an overview of the simulation of the interrupted knockout stages of Champions League $2019$-$20$.  Our model predicts both match(Total Goals, Total Passes, Possession, Blocks, Corners, etc.)  and player statistics(Who scored the goals, Assists, Shots, Crosses, etc.) for the two teams in the fixture. The winner(team with a higher aggregate score over two legs) proceeds to the next round.
In the case of a draw in the overall fixture (equal aggregate score from home/away legs), the team with the highest number of shots on target qualifies. We picked \textbf{Shots on target} as a decider, as it has the highest correlation with goals, which can be seen in Figure \ref{fig:corr}\subref{sub:cor}. 

The first simulation is between Bayern Munich and Chelsea(2nd Leg). Bayern Munich beat Chelsea comprehensively in the first leg fixture, which was conducted before the season was interrupted. Bayern entered the game with a form of five wins in its last five games, whereas Chelsea had mixed results recently. The odds favored Bayern to win this tie, which is also backed up by our results. Bayern beat Chelsea comfortably with a scoreline of $2$-$1$ dominating the possession($57\%$) and total passing($597$) stats. These stats are also backed up, as our data shows that Bayern Munich is one of the best teams in Europe in terms of passing and possession stats, which can be seen in Figure \ref{fig:stats}(1a) and Figure \ref{fig:stats}(2a). The goal scorers for Bayern were Robert Lewandowski and Jerome Boateng. Jorginho was the lone scorer for Chelsea. Our analysis shows Lewandowski as one of the most prolific goal scorers in Europe over the past few years, which is backed up by these results. 

A similar result was found in the simulation of the game between Barcelona and Napoli. Barcelona being European giants and one of the best passers in Europe dominated the passing($571$) and possession($56\%$) stats and won with a scoreline of $2-1$ at home with Rakitic scoring for Barcelona. Rakitic has a good record of scoring in Champions League knockouts, which is an interesting observation that our model is able to capture. Also, Barcelona has a great home record, as can be seen in Figure \ref{fig:dataset}, which is also corroborated by our results. 

In another match, Paris (PSG) beat Atlético by two goals to one in both fixtures. Our analysis shows that Paris, a team with a good scoring record (from Figure \ref{sub:cross}), have a tendency to perform better against more defensive teams like Atlético. Cavani, who is one of the most prolific scorers, scored in the fixture- thus validating our results. Another big fixture was the game between Juventus and Man. City in which Ronaldo scored one goal, and Dybala scored two goals. However, their efforts were in vain, as Laporte scored two headed goals off corners, and Gabriel Jesus scored one to take Manchester City to the semi-finals against Paris. Paris, being the in-form team in the semi-finals, beat Manchester City by dominating them in terms of both possession($58\%$) and passing stats, where Cavani and Peredes scored. This fixture at Manchester City's home ground was level in terms of possession($50\%$) and passing statistics which can be explained by Man. City's strong record at home, wherein they lost only $3$ out of $18$ games as seen in Figure \ref{fig:dataset}. These results validate our model's ability to learn about interactions between features.  

The other semi-final was a close fixture between Bayern Munich and Barcelona. Both teams, being two of the favorites, dominated the stats at home. They established a strong home record and the match ended in a draw, with Bayern decided as winners on the basis of the highest number of shots on target (as per our chosen method). Another interesting observation was that our model could not decide the winner in this fixture over both legs, which is expected since Bayern and Barcelona were favorites to win the competition.

The final was played between Bayern Munich and Paris, where Bayern Munich emerged victorious. Few exciting observations from this simulation are discussed as follows: Lewandowski scored two goals for Bayern Munich, making a substantial contribution to Bayern's success. Bayern Munich had the highest blocks per game in the simulations, which can be explained by Manuel Neuer's brilliant performances over the last few years. Finally, the results of our model are also backed up by the fact that Bayern Munich is one of the strongest teams in the competition, and had the best form leading up to the knockout stages. 

Our model can not only be used for predicting match statistics, but also for tactical analysis to help teams prepare better. We have shown that our model can make optimal predictions, and thus teams can use these predictions to be better prepared against their opposition. For example, in the simulation of a game between Bayern and Chelsea, our model predicted a significantly large number of crosses from Bayern, which matches their playing style and also reflects how a team is likely to play against another. Such analysis can help teams to plan better by focusing more on defending crosses if it is the opposition team's expected mode of attack. In addition, masked relations such as the performance of a team against relatively aggressive/defensive teams can be analysed and used to alter tactics accordingly. Finally, in order to verify the robustness of our model, we present some visualizations in Figure \ref{fig:stats}. We show the distributions of Passes, Possession, and Corners in the training data and their distributions in predictions of our simulation. It is seen that Barcelona and Bayern lead most of these stats in the training plots, and similar distributions can be seen in the simulations. It is evident from the plots in Figure \ref{fig:stats} that our model is robust and can capture the information and interactions among features very well.

\section{Conclusion and Future Work}

Inspired by the recent focus on sports analytics, and curiosity among the community on how the current seasons would have concluded, we conducted a simulation to find out how the rest of the season would pan out. We present \textbf{UCLData}, which contains data from the UCL games between the seasons $2014$-$2020$. We also propose a novel architecture that can efficiently capture the information and interactions within this data and make robust predictions on how individual matches of the season will pan out. We also propose solutions to handle some common problems related to data bias. Finally, we predict the results of the remaining Champions League games and thus predict the winners of this year's Champions League.

Future work can focus on giving weightage to the time of the matches, i.e. older matches will have a lower weightage as compared to the newer ones in the embedding. Although our model seems to work great on UCLData, it would be interesting to assess its learning capabilities on future football events and data from other leagues as well. Our methodology can be extended to predict other specific statistics such as the exact time of goals. Also, in the cases of a tied fixture over both legs, a penalty shootout simulation can also be added. In addition we would like to extend this work to more sporting events in the future.
\bibliography{main}

\begin{thebibliography}{10}

\bibitem{visual1}
Roman Voeikov, Nikolay Falaleev, and Ruslan Baikulov.
\newblock Ttnet: Real-time temporal and spatial video analysis of table tennis.
\newblock In {\em Proceedings of the IEEE/CVF Conference on Computer Vision and
  Pattern Recognition Workshops}, pages 884--885, 2020.

\bibitem{visual2}
H.~{Shih}.
\newblock A survey of content-aware video analysis for sports.
\newblock {\em IEEE Transactions on Circuits and Systems for Video Technology},
  28(5):1212--1231, 2018.

\bibitem{tactical}
Robert Rein and Daniel Memmert.
\newblock Big data and tactical analysis in elite soccer: future challenges and
  opportunities for sports science.
\newblock {\em SpringerPlus}, 5, 12 2016.

\bibitem{kampakis2015using}
Stylianos Kampakis and William Thomas.
\newblock Using machine learning to predict the outcome of english county
  twenty over cricket matches.
\newblock {\em arXiv preprint arXiv:1511.05837}, 2015.

\bibitem{geneticball}
Alexander~P Rotshtein, Morton Posner, and AB~Rakityanskaya.
\newblock Football predictions based on a fuzzy model with genetic and neural
  tuning.
\newblock {\em Cybernetics and Systems Analysis}, 41(4):619--630, 2005.

\bibitem{footbayesian}
Anito Joseph, Norman~E Fenton, and Martin Neil.
\newblock Predicting football results using bayesian nets and other machine
  learning techniques.
\newblock {\em Knowledge-Based Systems}, 19(7):544--553, 2006.

\bibitem{kickoffai}
Lucas Maystre and Victor Kristof.
\newblock Kickoff.ai uses machine learning to predict the results of football
  matches.

\bibitem{ieee34}
Anito Joseph, Norman~E Fenton, and Martin Neil.
\newblock Predicting football results using bayesian nets and other machine
  learning techniques.
\newblock {\em Knowledge-Based Systems}, 19(7):544--553, 2006.

\bibitem{gaussian}
Lucas Maystre, Victor Kristof, Antonio J~Gonz{\'a}lez Ferrer, and Matthias
  Grossglauser.
\newblock The player kernel: learning team strengths based on implicit player
  contributions.
\newblock {\em arXiv preprint arXiv:1609.01176}, 2016.

\bibitem{worldcup}
Kou-Yuan Huang and Wen-Lung Chang.
\newblock A neural network method for prediction of 2006 world cup football
  game.
\newblock In {\em The 2010 international joint conference on neural networks
  (IJCNN)}, pages 1--8. IEEE, 2010.

\bibitem{noconf}
Josip Hucaljuk and Alen Rakipovi{\'c}.
\newblock Predicting football scores using machine learning techniques.
\newblock In {\em 2011 Proceedings of the 34th International Convention MIPRO},
  pages 1623--1627. IEEE, 2011.

\bibitem{nn1}
S~Mohammad Arabzad, ME~Tayebi~Araghi, S~Sadi-Nezhad, and Nooshin Ghofrani.
\newblock Football match results prediction using artificial neural networks;
  the case of iran pro league.
\newblock {\em Journal of Applied Research on Industrial Engineering},
  1(3):159--179, 2014.

\bibitem{nn2}
Andrew~M Flitman and Enn~S Ong.
\newblock Using neural networks to predict afl game outcomes.
\newblock In {\em IEEE Conference on Computational Intelligence and Multimedia
  Applications}, pages 291--295. Griffith, 1997.

\bibitem{soup}
Leonard Richardson.
\newblock Beautiful soup documentation.
\newblock {\em April}, 2007.

\bibitem{fbref}
Sean Forman and Mike Kania.
\newblock Football statistics and history.
\newblock \url{https://fbref.com/en/}.

\bibitem{gsa}
Global sports archive.
\newblock \url{https://globalsportsarchive.com}.

\bibitem{autoencoder}
David~E Rumelhart, Geoffrey~E Hinton, and Ronald~J Williams.
\newblock Learning internal representations by error propagation.
\newblock Technical report, California Univ San Diego La Jolla Inst for
  Cognitive Science, 1985.

\end{thebibliography}

\end{document}